\documentclass[a4paper, 10pt, conference]{ieeeconf}      

\IEEEoverridecommandlockouts                              

\usepackage{amsmath} 
\usepackage{amssymb}  
\usepackage{bm}
\usepackage{graphicx}
\usepackage{epstopdf}
\usepackage{tikz}
\usepackage{pgfplots}
\usepackage{subcaption}
\usepackage{ulem}
\usepackage{tabularx}
\usepackage{comment}
\usepackage{placeins}
\usepackage{gensymb}

\title{\LARGE \bf 
Kinematic Modeling of Handed Shearing Auxetics via Piecewise Constant Curvature}

\author{Aman Garg$^{1}$, Ian Good$^{1}$, Daniel Revier$^{1,2}$, Kevin Airis$^{1}$, and Jeffrey Lipton$^{1,2}$\\
\thanks{This work was supported by the NSF grant number 2035717, Murdock Grant 201913596, and ONR grant DB2240}
\thanks{$^{1}$ Mechanical Engineering Department, The University of Washington, Seattle, WA, 98195 USA}
\thanks{$^{2}$ Paul G. Allen School of Computer Science and Engineering, The University of Washington, Seattle, WA 98195}
}

\usepackage{xcolor}

\begin{document}

\maketitle
\thispagestyle{empty}
\pagestyle{empty}

\begin{abstract}




Handed Shearing Auxetics (HSA) are a promising technique for making motor-driven, soft, continuum robots.  Many potential applications from inspection tasks to solar tracking require accurate kinematic models to predict the position and orientation of these structures. Currently there are no models for HSA based continuum platforms. To address this gap we propose to adapt Piecewise Constant Curvature (PCC) Models using a length change coupling matrix. This models the interaction of HSA structures in a 2x2 array. The coupling matrix maps the change in motor angles to length changes and defines the configuration space in our modified PCC Model. We evaluate our model on a composite movement encompassing bending, extension and compression behavior. Our model achieves a positional accuracy with mean error of 5.5mm or 4.5\% body length and standard deviation of 1.72mm. Further, we achieve an angular accuracy with mean error of  -2.8$^\circ$ and standard deviation of 1.9$^\circ$. 

\end{abstract}

\section{Introduction}


The ability to orient robotic arms in space is critical to a wide range of applications. In inspection tools the ability to point cameras and other sensors ensure our critical infrastructure is well maintained and objects we purchase are fabricated properly. Satellite communication systems require accurate angular tracking to ensure connections are not lost~\cite{sattelite5342320}\cite{innovationsSatelliteCommunications}. In energy systems it enables sun tracking for solar panels and solar thermal systems\cite{MOUSAZADEH20091800}\cite{ROTH2004393}. All of these applications necessitate that they are resilient to contact with their environments which is a strength of soft robotics\cite{SoftRoboticsReviewhttps://doi.org/10.1002/adem.201700016} \cite{resulience10.3389/frobt.2017.00048} \cite{Rus2015}. An electrically driven soft robot able to accurately point would be of broad utility.

Current research on electrically driven soft actuators has shown that handed shearing auxetics (HSAs) have proven very promising \cite{Lipton632}. They are capable of linear expansion, and have distributed compliance like pneunets\cite{mosadegh2014pneumatic}, but are driven by motor torques rather than fluid flows.  Recent research has focused on applying HSAs to gripping tasks\cite{Chin20198794098,chin2020multiplexed}, sensorization\cite{chin2019simple,chin2019automated} and  mechanical characterization\cite{Truby2021-9326362}, but there have been no proposed control methods that describe the actuation of HSAs. 

\begin{figure}[t]
    \centering
    \hfill
    \centering
    \includegraphics[width=1\linewidth]{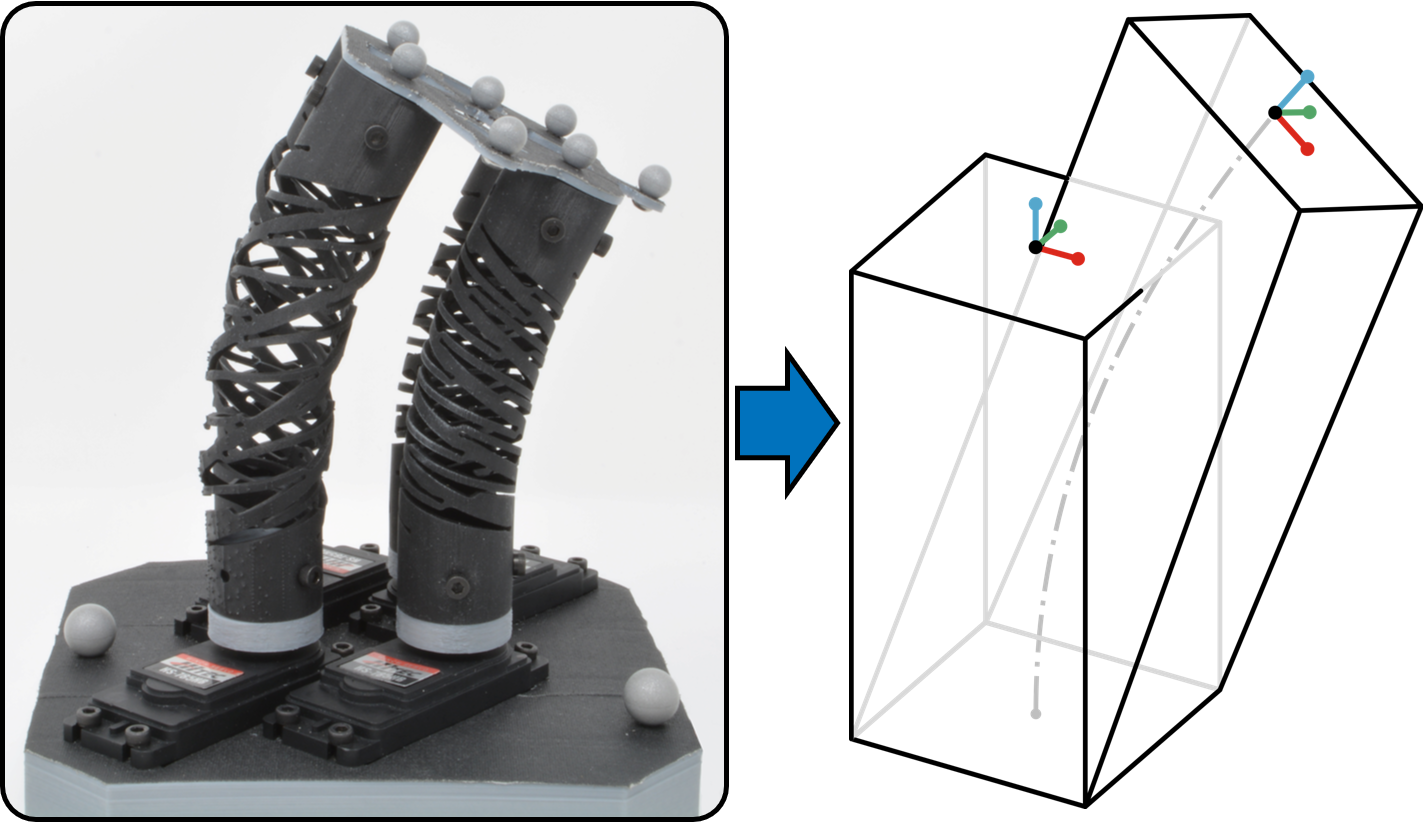}
    \captionsetup{justification=centering}
    \caption{Modeling a 2x2 HSA platform with Extended Piecewise Constant Curvature}
    \label{fig:hero}
    \hfill
    \vspace{-0.75cm}
\end{figure}

We use a common design of two sets of opposite handed HSAs to form a 2x2 HSA platform\cite{Lipton632} as seen in Fig.\ref{fig:hero}. 
Its  simple design makes it an ideal section for a soft robotic inspection arm, while its large platform area makes a single section ideal for used as a pointer for solar panels or communication equipment. Despite the simplicity and utility of the design, without a model it cannot be of use or properly controlled. 

We therefore look to adapt the existing soft robotics literature on control and apply them to HSA systems in order to enable electric soft robotic pointing systems and future multi-segment arms. We look to a common soft robot modeling method,  Piecewise Constant Curvature (PCC)\cite{Walker2006-1588999,Jones2010-doi:10.1177/0278364910368147}. PCC models assume that arms can be modeled as composites of rods that bend at a programmable radius and length that rotate relative to each other. Typically this bending is induced by differential pressure on chambers\cite{katzschmann2019dynamic}. Because HSAs are extensible and compliant, we needed to develop a constant curvature model that captures this behavior. Extensible PCC models capture the arc length changes of the segments as they are actuated.

Our work maps the actuation space to segment lengths and accounts for inter-segment coupling with 5 parameters in a coupling matrix. We demonstrate that the PCC is more predictive than simple kinematic models with and without coupling. We evaluate this model on a  composite movement to combine bending, extension and compression of the HSAs. Our proposed model achieves a positional accuracy with mean error of 5.5mm or 4.5\% body length and standard deviation of 1.72mm. For the orientation, we achieve an angular accuracy with mean error of  -2.8$^\circ$ and standard deviation of 1.9$^\circ$.

In this paper we:
\begin{itemize}
    \item Model the interactions between individual HSAs on a platform.
    \item Adapt the Piecewise Constant Curvature model to a HSA driven platform with 4 lengths.
    \item Validate the modified PCC model through experiments on a HSA platform.
\end{itemize}

\section{Background and Related Work}

Soft robots have been usually driven by variable length tendons and pneumatic actuation and more recently, electroactive polymers \cite{Shintakeelectroactive}.Handed shearing auxetics (HSAs) are a new category of material that enables electric soft robots. HSAs couple auxetic expansion with shear as a result of their periodic links and joints structure \cite{Lipton632}. Auxetic metamaterials normally have a symmetric point in their trajectory, enabling them to transition from one handedness to another. This symmetry is purposely broken in HSAs, resulting in a single chiral shear motion. In order to create a HSA of opposite handedness, a simple reflection of the periodic pattern is sufficient~\cite{Truby2021-9326362}. When tiled on a cylinder, the HSA patterns couple global shearing from twisting to expansion along the axis of rotation~\cite{Chin20188404904}. These structures are compliant by nature, allowing for spring like responses to impacts or other disturbances. Their responsiveness to twisting allows for them to form linear compliant actuators driven directly by motor torques when placed in pairs~\cite{Lipton632}. Most applications of HSAs have been in grippers, since they require no model to be able to grasp successfully~\cite{Chin20198794098,chin2020multiplexed,chin2019simple,chin2019automated}. 

Modelling the motion of soft robots is challenging due to their inherent infinite degrees of freedom.Various research works have shown that finite element methods and other numerical techniques can be used to describe the behaviour{\cite{zhang2016kinematic}}{\cite{elsayed2014finite}}. But their inability to provide a closed form solution limits their adoption. An alternate approach - piecewise constant curvature models have been found to be more effective in defining the behaviour, where the robotic structure is broken into continuous segments with constant curvatures providing an easy way to model a system which is differentiable everywhere{\cite{8722799}}{\cite{webster2010design}}. PCC modelling includes two mappings - 1) From actuator space to the configuration space ($s$,$\kappa$,$\phi$) which describes the curvature produced by the continuum robot and 2) from configuration space to pose which defines the position and orientation of the end-effector. 

The first mapping is unique to the type of robotic system such as pneumatic or tendon actuated arms{\cite{renda2014dynamic}}{\cite{talas2020design}}. The second mapping is a standard kinematic transformation from the arc parameters to the coordinate system of the task space. This transformation can be achieved through different conventions and definitions{\cite{Walker2003-https://doi.org/10.1002/rob.10070}}. One of those techniques uses the modified Denavit-Hartenberg (D-H) approach to define a transformation matrix that maps the configuration space to the task space.In this modified D-H method, the continuum robot is defined as a virtual rigid-link robot allowing for convenient closed form calculations \cite{Walker2006-1588999}.    

On first glance, one might be inclined to use a parallel manipulator model for the kinematics of HSA based robotic section, since both HSAs and Stewart platforms rely on a change of length\cite{dasgupta2000stewart}. However, unlike links in a Stewart platform HSAs are always perpendicular to their base. This is most similar to traditional pneunet based structures which have been modeled using PCC \cite{della2020data}. 



\begin{figure}
    \centering
    \begin{subfigure}[t]{0.5\textwidth}
        \centering
        \includegraphics{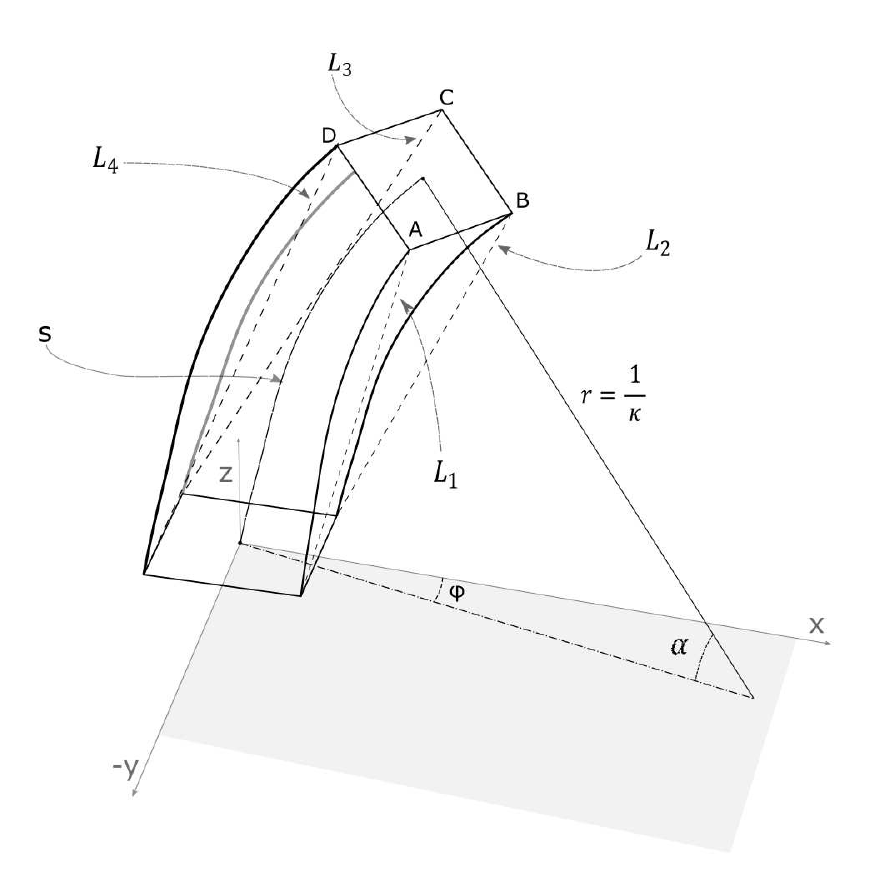}
        \caption{Configuration parameters for the PCC model}
        \label{fig:configuration}
    \end{subfigure}
    \hfill
    \begin{subfigure}[t]{0.15\textwidth}
        \centering
        \includegraphics[width=1\textwidth]{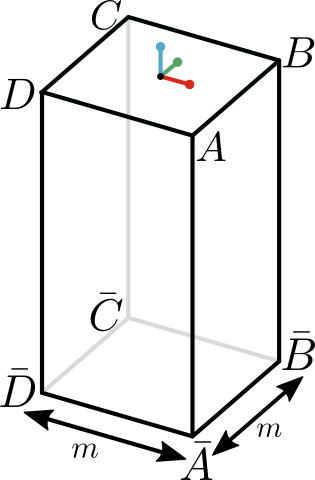}
        \caption{Assembly at rest.}
        \label{fig:rest_position}
    \end{subfigure}
    \hfill
    \begin{subfigure}[t]{0.21\textwidth}
        \centering
        \includegraphics[width=1\textwidth]{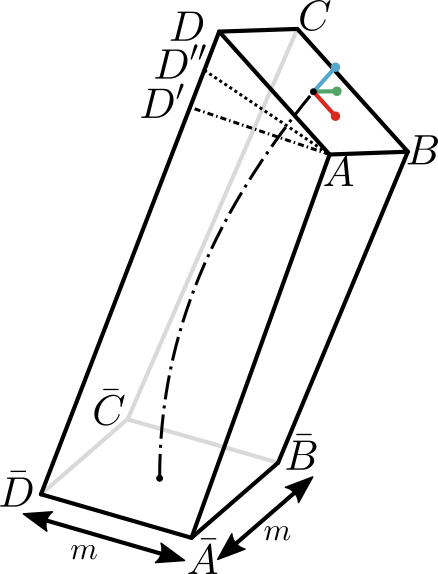}
        \caption{Assembly after actuation.}
        \label{fig:actuated_position}
    \end{subfigure}
    \hfill
    \caption{The piecewise constant curvature model maps the location of a robot section using curvature($\kappa$) arclength ($S$) and }
    \label{fig:configurationFull}
\end{figure}

\section{HSA model and Platform Design}
\label{sec:HSADesign}
HSAs, though not conventional linear actuators, do extend when the connected motor is activated, so the extension of a single HSA is a function of the motor's angle of rotation $\theta$. We made our HSAs on a Carbon M1 out of FPU50 with a inner diameter of 22.98 mm and thickness of 2 mm. 
To establish mapping between angles and HSA length we conducted a force vs. displacement test using an Instron Universal Testing Machine at step angles of 30\textsuperscript{o} from 0\textsuperscript{o} to 180\textsuperscript{o}. At each step we observed the length at which the force was zero. We observed that the extension is approximately linearly dependent on the angle and as a result the length of an actuator can be modeled as linear function of the angle of the servo motor.  We found an equation of the form $L = \alpha_{0}\ast \theta   + \alpha_{1}$ with $\alpha_{0}=$ 0.124, $\alpha_{1}=$ 0.119 fit the data with an $R^2=$ 0.99. 

%
%

\subsection{The 2x2 Platform}

\begin{figure}[htbp]
    \centering
    \begin{frame}{ 
    \includegraphics[width=.7\linewidth]{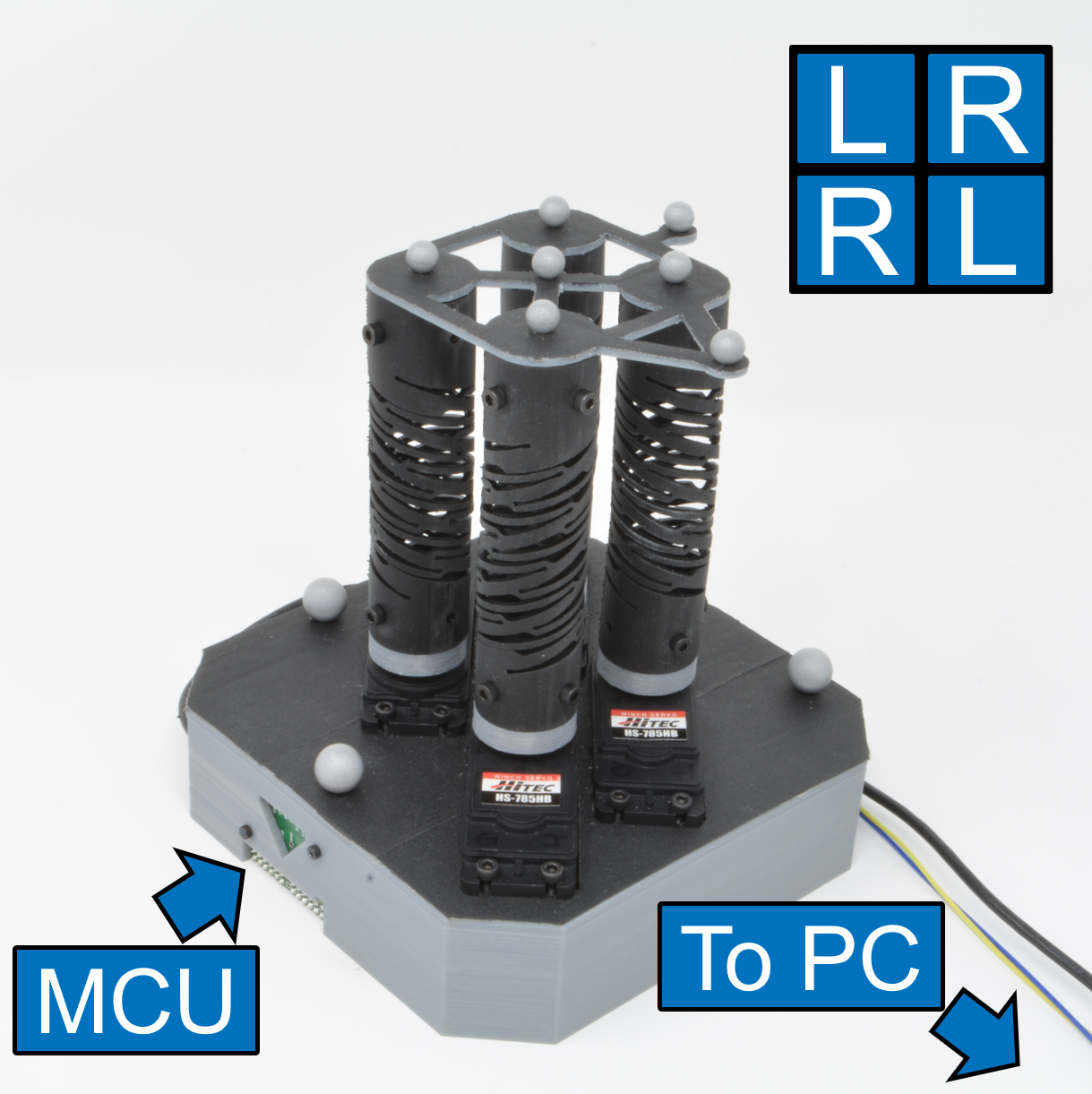}}    
    \end{frame}
    \captionsetup{justification=centering}
    \caption{The 2x2 HSA test platform. The top figure indicates the handedness of each HSA in the platform.}
    \label{fig:HSA_Platform}
\end{figure}

\begin{figure}
    \centering
    \includegraphics{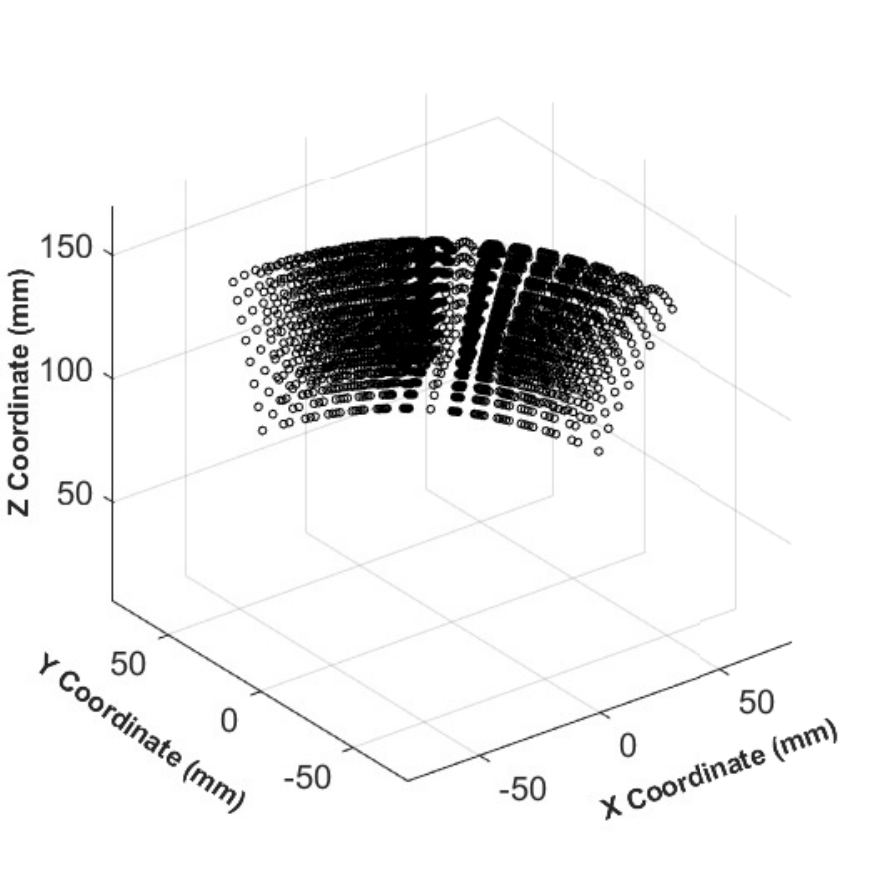}
    \captionsetup{justification=centering}
    \caption{Workspace of the HSA platform }
    \label{fig:workspace}
\end{figure}

A 2x2 HSA platform Fig.~\ref{fig:HSA_Platform} was designed, 3D printed, and assembled. It has four independent HS-785HB servos in the base connected to to four vertical HSAs. The HSAs are attached together to a ridged platform on top. The servos are controlled via a Pololu Mini Maestro 12-Channel USB servo controller and are powered with 6V and the system draws up to 3 watts during testing. The servos are mounted to the base in an angled pattern at 42$^{\circ}$. This arrangement allows for a 43.68 mm center to center distance between HSAs. A key feature of the actuator assembly is the presence of opposite handed HSAs. In order to balance the net torque on the top platform produced from the independent rotation of each HSA, each handed pair of actuators are arranged diagonally in the 2x2 grid as shown in the top right corner of Fig.\ref{fig:HSA_Platform}. Thus, two of the servos rotate clockwise and two rotate counterclockwise to extend each HSA. Therefore there is no need for a strain constraint layer or a solid core in the structure. This allows the platform to change in length and extend in the Z direction in addition to bending. The resulting workspace of a single structure can be seen in Fig~\ref{fig:workspace}.

\section{Kinematic Models}

%

\subsection{Mapping $\theta$ to Length}

In the 2x2 HSA platform, all the actuators are fixed in position at the base and         coupled together with the top platform acting as a second common link between them. If HSA actuators were rigid length change actuators, the system would appear to be over constrained since only 3 heights relative to one plane are needed to define another. HSAs however are compliant, with the ability to compress and extend in response to external forces in addition to twist driven length changes. This means that the length of any of the actuators at any time is dependent on length of other actuators (and by extension their theta states). We use our previous observation to model the length of actuators as linearly dependent on theta values of all actuators. If $\vec{L}$ represents the lengths of the actuators then we can write $\vec{L}$ as
\begin{equation*}
    \vec{L} = \bar{\beta_0} \cdot \vec{\theta}   + \vec{\beta_{1}}.    
\end{equation*}

The $\beta_0$ matrix establishes the extent of coupling between the actuators and $\vec{\theta}$ and $\beta_1$ vector establishes their height at rest. Consequently, any change in the dimensions of HSA would affect the coupling. For this reason we determine $\beta_0$ and $\beta_1$ for our chosen design through regression modelling.
 
The $\beta_\text{0,regression}$ matrix does not result in a perfectly symmetric matrix. To enforce symmetry we break the $\beta_0$ matrix up into two components, $\beta_\text{0,symmetric}$ and $\beta_\text{0,deviation}$ such that $\beta_\text{0,symmetric}$ + $\beta_\text{0,deviation}$ = $\beta_\text{0,regression}$. The $\beta_\text{0,symmetric}$  matrix is then found by taking the mean of the diagonals, making a circulant matrix. We then use $\beta_\text{0,symmetric}$ as $\beta_0$. Further, we average out the terms in $\beta_1$ to build a measure of the height of the actuators at rest.

\subsection{Piecewise Constant Curvature approximation of HSAs}


We choose the central point between all HSAs to be the distal end of our constant curvature rod as seen in Fig.~\ref{fig:actuated_position}. From there, we define the origin to be the point between the bottom of all the HSAs.
In order to model the HSA assembly we first establish the straight lines connecting the ends of the HSAs. Fig.~\ref{fig:rest_position} shows the HSA assembly represented as two square, rigid bodies of the same dimension $m$, with corners $P = \{A, B, C, D\}$ and $\bar{P}=\{\bar{A},\bar{B},\bar{C},\bar{D}\}$ called the ``platform'' and ``base'' respectively. We also initially constrain all points in $P$ to lie in the same plane $\mathcal{P}$ and all points in $\bar P$ to lie in the plane $\mathcal{B}$. The platform and base are separated by the set of rigid links with lengths $L=\{l_A, l_B, l_C, l_D\}$ where $l_J=\|\bar{J}J\|_2$ for $J \in P$ and have geometric centers. We note that due to this geometry the distance ${l_S}$ from the center of the platform to the origin i.e. the center of the base is given by: 

\begin{equation} \label{eq:ls}
    l_S = \frac{l_A + l_C}{2} = \frac{l_B + l_D}{2} 
\end{equation}

Though we can calculate these lengths for each individual actuator, more significant is their  arc length subtended during it's motion. Given the dependence of the platform's pose on angles and curvatures subtended by actuators, we follow the piecewise and extensible constant curvature approach to model the position and orientation as a function of their lengths.


\subsection{Calculation of $\kappa$, $\phi$, and $s$}

To determine $\kappa$, $\phi$, and $s$ for our system, we extend the previous work on PCC models and determine the following relationships while referring to Figure~\ref{fig:configurationFull}.

\begin{equation}
\begin{aligned}
    \kappa 
          &= \frac{1}{d} \sqrt{\frac{(l_C-l_A)^2}{(l_C + l_A)^2} + \frac{(l_D-l_B)^2}{(l_D+l_B)^2}}
\end{aligned}
\end{equation}

The angle of rotation about the z-axis $\phi$ is derived as

\begin{equation}
\begin{aligned}
    \phi  
         &= tan^{-1}\left(\frac{l_Bl_C - l_Al_D}{l_Cl_D-l_Al_B}\right)
\end{aligned}
\end{equation}



    

Finally the arc length $s$ is proportional to the angle $\phi$ by the curvature $\kappa$, 
\begin{equation}
    s = \frac{2}{\kappa} sin^{-1}\left(\kappa\frac{l_S}{2}\right).
\end{equation}
It is worth noting that
\begin{align*}
\lim_{\kappa\to 0} s(\kappa, l_S) &= l_S \\
\lim_{\kappa_x \to 0^{\pm}} \phi(\kappa_y, \kappa_x) &= \pm \frac{\pi}{2}\
\end{align*}
which are the cases for when the platform is in its rest position and when there is curvature only in the $YZ$ plane respectively. 



\subsection{Calculation of position and orientation}

With the center of the top platform represented in arc parameters ($s,\kappa , \phi$), we use a D-H formulation for transformation to X ,Y, Z coordinates of the center of the top platform. We approach the forward kinematics from the base as a virtual revolute-revolute-prismatic (RRP) transformation as previously stated in \cite{Jones2010-doi:10.1177/0278364910368147}. This approach defines our transformation matrix shown in \eqref{eq:T}.
\begin{equation}
    T = \begin{bmatrix} \cos\phi\cos\kappa s & -\sin\phi & \cos\phi\sin\kappa s & \frac{\cos\phi(1-\cos\kappa s)}{\kappa} \\ \sin\phi\cos\kappa s & \cos\phi & \sin\phi\sin\kappa s & \frac{\sin\phi(1-\cos\kappa s)}{\kappa} \\ -\sin\kappa s & 0 & \cos\kappa s & \frac{\sin \kappa s}{\kappa} \\ 0 & 0 & 0 & 1 \end{bmatrix}
    \label{eq:T}
\end{equation}
The base coordinates of the system are defined in our system as - 
\begin{equation}
   B = [0 ; 0 ; 0; 1] 
   \label{eq:basezero}
\end{equation}

The position coordinates of the center point of the top platform would then be given by:
\begin{equation}
P = \begin{bmatrix}
x \\
y \\
z \\
1
\end{bmatrix}
= T\ast B
\label{eq:Pcoordinate}
\end{equation}

Further, the orientation is defined by two angles:

\begin{equation}
  \begin{bmatrix}
\alpha \\
\phi \\

\end{bmatrix}
= \begin{bmatrix}
s\kappa \\
tan^{-1}(\frac{l_Bl_C - l_Al_D}{l_Cl_D-l_Al_B}) \\

\end{bmatrix}
\label{eq:orientationvalue}
\end{equation}

\begin{figure*}[thbp]
    \centering
    \includegraphics[width= 17cm]{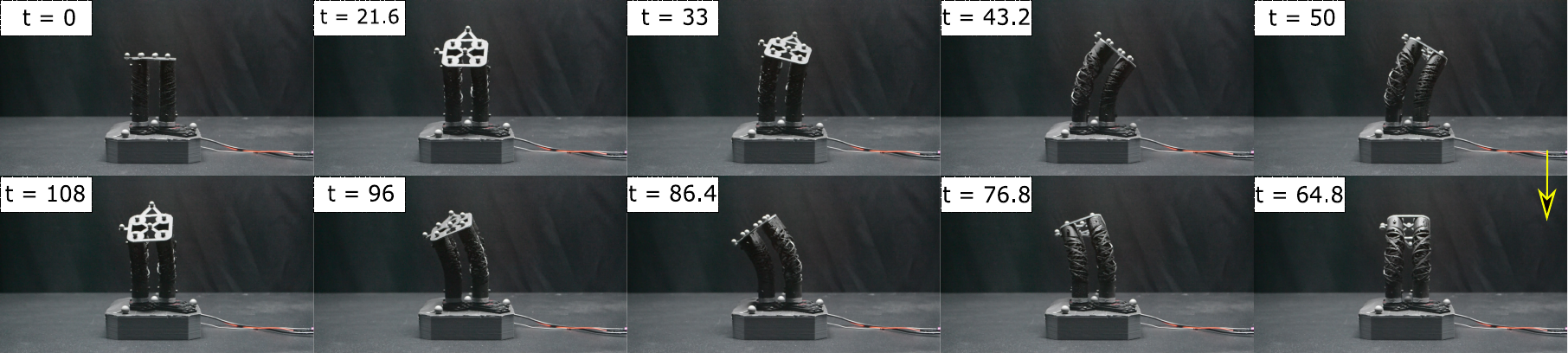}

    \caption{Sequence of operations with the platform for experimental validation where each frame represents the state of the platform at time = t seconds.}
    \label{fig:sequence}
\end{figure*}

\section{Experimental Setup}
Calibration is required to map between PWM values in Pololu's software to angular displacement on each individual servo for precise and accurate control. This was done by attaching an approximately 10in rod to the end of each servo and rotating 90 degrees to determine a conversion between PWM values and servo's angle of rotation.An OptiTrack system made from four Flex 13 cameras was used to track changes in the 2x2 HSA platform. The four cameras were located in top corners of a 1 cubic meter test volume and pointed at the platform. Two rigid bodies were defined representing the dynamic top platform and the static bottom platform. 3D position and quaternion data were saved for each rigid body. We gathered all permutations on the HSAs from 0$^{\circ}$ to 180$^{\circ}$ in 30$^{\circ}$ increments. This resulted in 2401 unique states. These states were used to determine the coupling matrix ${\beta_0}$ and vector ${\beta_1}$. 

For validation , we performed a composite rotation of the platform as shown in Fig~\ref{fig:sequence}.This ensured that all the possible positions - extension, compression, bending \& rotation were combined within one single sweep. We performed this sweep for the full range of speeds the servos could generate (between 0.15 RPM and 30 RPM)

\section{Results and Discussion}
  


\subsection{Mapping $\theta$ to length}
\begin{figure}
    \centering
    \includegraphics{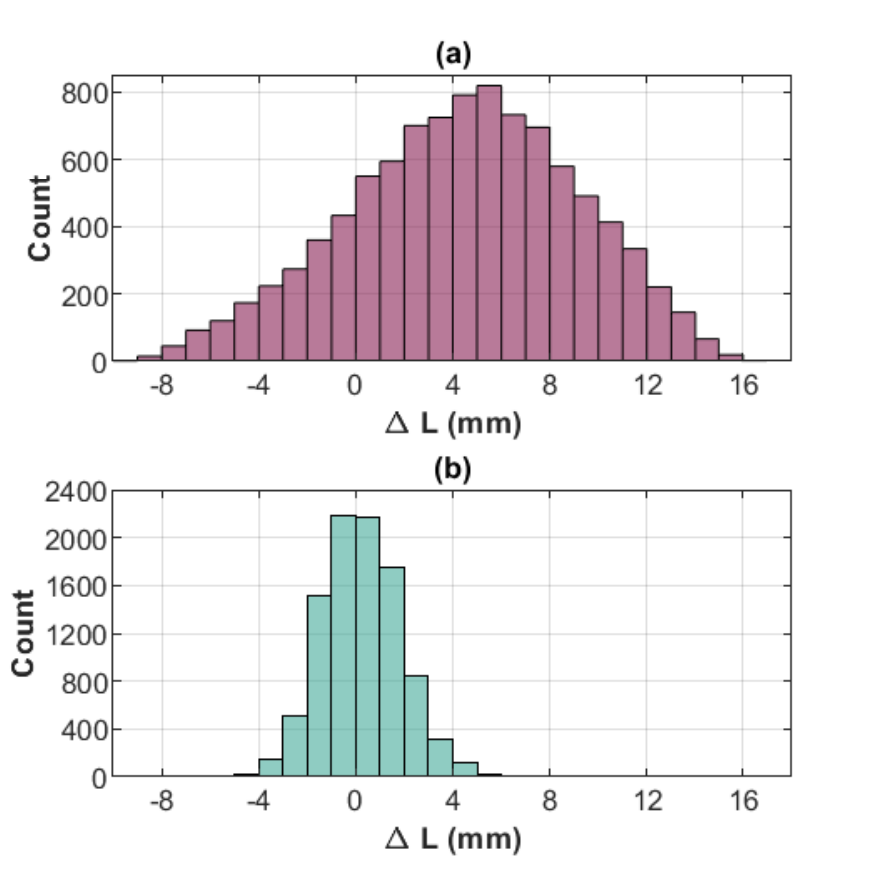}
    \captionsetup{justification=centering}
    \caption{(a) Error between the estimated and observed actuator length when using uncoupled mapping ; (b) Error between estimated and observed actuator length when using coupled mapping }
    \label{fig:coupled_errorl}
\end{figure}

The basis of our model is the mapping between inputs $\vec{\theta}$ and lengths. In section~\ref{sec:HSADesign} we found the linear relationship between $\theta$ and the length $l$ of an HSA at which there is no more extensional force. As a first approximation, we use these results in an uncoupled linear model for estimating lengths of the actuators and observe significant errors against measured lengths from OptiTrack. The analysis for it is shown in Fig.~\ref{fig:coupled_errorl}A where the average error in length estimation is 4.78mm with a standard deviation of 4.7mm and a range of -9 to 16mm. This was not unexpected as the above model considers the actuators as independent structures but the 2x2 HSA platform acts as a common linkage between the four actuators producing a coupled model with interconnected links. This leads the naive model to over estimate the length of the segments and produces a wide range of errors. 

\begin{figure}
    \centering
    \includegraphics{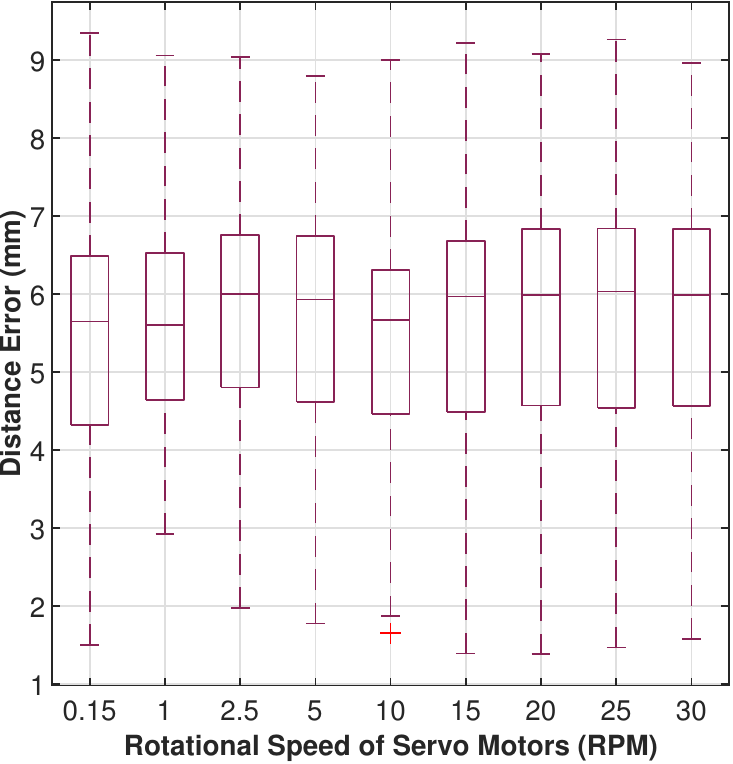}
    \captionsetup{justification=centering}
    \caption{Position error with change in velocity }
    \label{fig:position_error}
\end{figure}

By contrast when we integrate the two observations in our Coupled Length Estimation Model, as explained in the modeling section, we correct the bias and reduce the range and variance of the errors as seen in Figure~\ref{fig:coupled_errorl}. The resulting error in actuator length prediction reduces to a mean of 0.25mm with a standard deviation of 1.6mm with a range of -4 to 6 mm. Since the errors in length propagate through the model we must examine the effects on predicting position and state through the trajectory.

In Figure~\ref{fig:comparison}A we see the inputs to the sequence sequence seen in Fig~\ref{fig:sequence}. The lengths of the actuators predicted and measured are seen in Figure~\ref{fig:comparison}C.  As we can see, when the actuator is moving, we see the most agreement between the model and measurement. The greatest disagreement occurs when an actuator is meant to be held still. This is likely a result of an relationship not captured in the beta matrix. We suspect that the driving actuators, acting as a force generator are compressing the held actuators, which then are pulled or pushed. An alternative explanation where the limits of extension are the source of errors is less likely to be true since L2 and L1 have large errors at their rest state.



\subsection{Piecewise Constant Curvature Model}
\begin{figure*}[!th]
    \centering
    \includegraphics[width= 17cm]{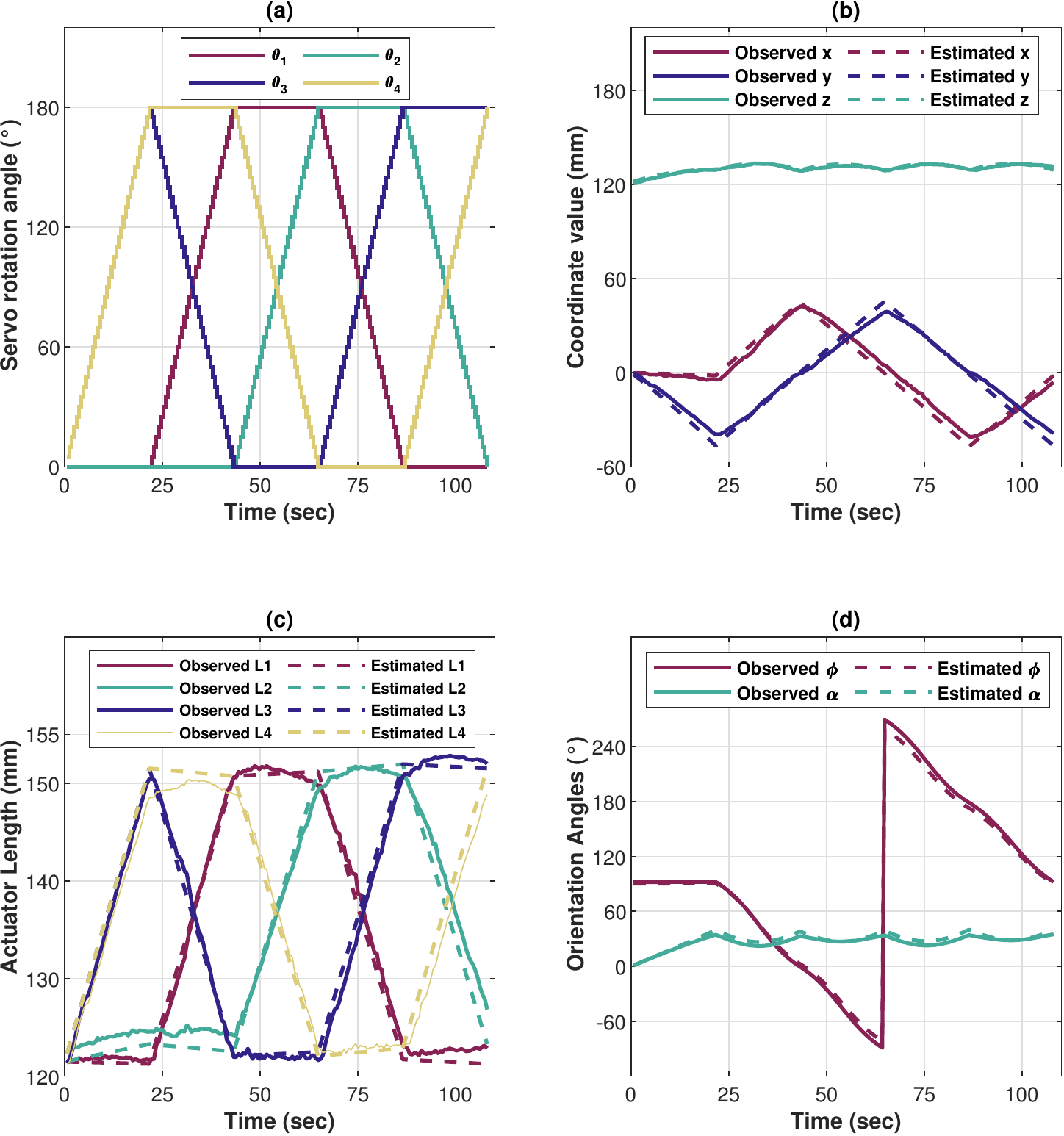}
    \captionsetup{justification=centering}
    \caption{(a) Shows the variation of servo motor angle(input) with time ; (b) Shows comparison of Observed vs Estimated position of the center of the top platform ; (c) Shows a comparison of Observed vs Estimated lengths of the actuators ; (d) Shows a comparison of Observed vs Estimated orientation angles of the platform. }
    \label{fig:comparison}
\end{figure*}

Figure~\ref{fig:comparison}B and D show the error propagation through the model. We cannot observe the full state of the PCC section model, but we can observe the value of $\phi$ from the Optitrack data since its merely the angle between the plane that contains the center point and the Z axis, with the XZ plane. In Figure~\ref{fig:comparison}D we see that $\phi$ is accurately captured by the model. The error of $\phi$ is between -3.4$^{\circ}$ and 3.7$^{\circ}$ with an average of -1.5$^{\circ}$. 

We can also observe $\alpha$, the angle subtended by the arc section which is a combined metric of $\kappa$ and $s$. $\alpha$ and $\phi$ encode the orientation of the platform.
We observe that $\kappa$ is well predicted from the kinematic model with a range of error between -5.85$^{\circ}$ and 0.2$^{\circ}$ and an average error of -2.8$^{\circ}$.
Therefore the model can reasonably capture the orientation of the platform. 

We compare the Cartesian coordinate of the center point with the one predicted from the model in Figure~\ref{fig:comparison}B. The result is a model which can predict the position with minimal error. When we repeat the motion at various speeds as seen in Figure ~\ref{fig:position_error} we see that for these motors, the model does not have an appreciable change in error. The figure shows the distribution of errors in the distance between the center of the platform observed and predicted. As the speed varies the mean error goes from a minimum of 5.5mm to a maximum of 5.7mm with no meaningful correlation with speed. Therefore we can conclude that for these motors and mass, the inertial effects can be ignored and a kinematic model such as ours is sufficient.

\section{Conclusion and Future Work}

In this paper we established that Piecewise Constant Curvature models (PCC) are good forward kinematic models for a 2x2 handed shearing auxetic platform when interaction between HSAs is modeled. This is the first model of HSA platforms and shows that soft robotics modeling techniques can be adapted to HSA systems.  In comparison with a naive approach,  a model with coupling is found to be more accurate for predicting the lengths of the actuators. These can then be mapped into position and orientation of the HSA assembly's top platform. Over the range of Servo motor actuation, this model shows no degradation with increased speed.

Now that HSA platforms can be modeled they can be used more fully in robotic applications beyond gripping. Future work includes integrating our model with a controls framework for specific applications. Additionally, the platforms we have designed have only constituted a single link in the PCC framework. Extending this to a multi segment robotic arm made from HSAs is a a logical next step.

\bibliographystyle{IEEEtran}
\bibliography{IEEEabrv,6EPCC}

\end{document}